\pdfoutput=1

\documentclass[11pt]{article}

\usepackage{acl}

\usepackage{times}
\usepackage{latexsym}

\usepackage[T1]{fontenc}

\usepackage[utf8]{inputenc}

\usepackage{microtype}

\usepackage{inconsolata}
\usepackage{amsmath}
\usepackage{bm}
\usepackage{algorithm}
\usepackage{algpseudocode}
\usepackage{graphicx}
\usepackage{svg}
\usepackage{multirow}
\usepackage{booktabs}
\usepackage{tcolorbox}
\usepackage{xspace}
\usepackage{CJKutf8}
\usepackage{tabularx}
\usepackage{stfloats}

\newcommand{\method}{G-DIG\xspace}
\newcommand{\methodq}{G-DIG {w/o} Diversity\xspace}

\newtheorem{lemma}{Lemma}

\title{G-DIG: Towards Gradient-based Diverse and High-quality Instruction Data Selection for Machine Translation}

\author{Xingyuan Pan\footnotemark[2]~~~~Luyang Huang~~~~Liyan Kang\footnotemark[2]~~~~\textbf{Zhicheng Liu}\\\textbf{Yu Lu}~~~~\textbf{Shanbo Cheng}\footnotemark[3]\\
   ByteDance Research\\
\texttt{xypan00@gmail.com}~~~~\texttt{\{huangluyang,~chengshanbo\}@bytedance.com}}

\begin{document}
\maketitle

\renewcommand{\thefootnote}{\fnsymbol{footnote}}
\footnotetext[2]{The work was done when the author was an intern at ByteDance.}
\footnotetext[3]{Corresponding author.}
\renewcommand{\thefootnote}{\arabic{footnote}}

\begin{abstract}
Large Language Models (LLMs) have demonstrated remarkable abilities in general scenarios. Instruction finetuning empowers them to align with humans in various tasks. Nevertheless, the \emph{Diversity} and \emph{Quality} of the instruction data remain two main challenges for instruction finetuning. With regard to this, in this paper, we propose a novel gradient-based method to automatically select high-quality and diverse instruction finetuning data for machine translation. Our key innovation centers around analyzing how individual training examples influence the model during training. Specifically, we select training examples that exert beneficial influences on the model as high-quality ones by means of Influence Function plus a small high-quality seed dataset. Moreover, to enhance the diversity of the training data we maximize the variety of influences they have on the model by clustering on their gradients and resampling.
Extensive experiments on WMT22 and FLORES translation tasks demonstrate the superiority of our methods, and in-depth analysis further validates their effectiveness and generalization.\footnote{Code is available at \url{https://github.com/xypan0/G-DIG}}

\end{abstract}

  




\section{Introduction}
Large Language Models (LLM) have revolutionized the field of Natural Language Processing with their strong abilities in general language understanding and generation~\cite{ChatGPT,achiam2023gpt}. To enable this strong ability, instruction finetuning has been proposed to better align language models~\cite{wei2021finetuned,chung2022scaling,ouyang2022training}. Significant progress has been made on collecting extensive instruction finetuning data for better aligning LLMs to produce helpful responses~\cite{chung2022scaling}. 

We argue that \textit{Diversity} and \textit{Quality} of the instruction data present a pair of challenges for instruction finetuning.~\newcite{zhou2023lima} have demonstrated that a model trained with a limited, carefully curated dataset composed of high-quality and diverse examples outperforms models trained on larger, more extensive datasets during instruction finetuning. Subsequently, various methods have been proposed to automatically select high-quality and diverse training data from the large pool of instruction finetuning datasets~\cite{chen2023alpagasus,cao2023instruction}. Yet, these methods often rely on another model to decide quality or diversity, neglecting the inherent model behavior and strong ability of the LLM itself.



To this end, we propose \textit{G-DIG}, a novel gradient-based method to automatically select diverse and high-quality instruction finetuning data for machine translation. We use influence function~\cite{koh2017understanding}, a gradient-based method that quantifies the impact made by individual training samples. Concretely, we (1) measure the response quality of each training sample with the influence score of the training sample on test instances and (2) enhance the diversity of the training data by maximizing the variety of influences they have on the model.

Specifically, we hypothesize that high-quality data should have positive influences on high-quality test samples. Hence, we first manually create a small set of high-quality seed data and then automatically select high-quality data that have positive influences on seed data. Moreover, we utilize K-means clustering algorithms to cluster training data with similar influences, using gradients representing their influences on the model.
Unlike existing methods that introduce an external model to decide quality and diversity, our methods directly use model gradients, which capture the model behavior through learning algorithms and back to the training data.

We conduct experiments on Zh $\Rightarrow$ En and De $\Rightarrow$ En translation tasks. Specially, We collect large candidate pools and manually construct two small sets of seed data. We then finetune different LLM backbones on various sizes of selected subsets and compare their performances with different selective methods and existing SOTA LLMs. Under a thorough comparison in a range from 1k to 64k selected samples, our proposed method not only surpasses the baseline selective methods but also achieves competitive performance against SOTA models. Extensive experiments and in-depth analysis emphasize the need for data selection and demonstrate the effectiveness and generalization of our proposed methods.

\section{Related Work}
\paragraph{LLM for Machine Translation.} 
Due to their strong in-context learning and instruction-following abilities, powerful LLMs like GPT-4 have achieved remarkable progress in machine translation, with comparable performance to the top systems on the WMT translation task \citep{zhu2023multilingual,he2023exploring,raunak2023leveraging}. To fully leverage LLMs' translation ability, various methods have been proposed, including in-context translation exemplar selection \citep{garcia2023unreasonable,lin2022few,zhang2023prompting,agrawal2022context}, prompt optimization \citep{jiao2023chatgpt} and decoding strategies \citep{zeng2023improving}.

The aforementioned studies all focus on the inference stage optimization, while another line of work focuses on instruction tuning the LLMs for better translation performance. \citet{xu2023paradigm} proposes to first finetune the model on monolingual data and then on high-quality parallel data. \citet{li2024eliciting} trains the model to elicit translation ability by multilingual instruction tuning. \citet{li2024mt} proposes to create high-quality instruction-tuning data from larger models by a patching mechanism. \citet{chen2023improving} improves the model instruction understanding by adding a global instruction representation and improves model faithfulness by comparing over-translation and misstranslation results with the correct translation. \citet{zeng2023tim} proposes a novel framework using examples in comparison to teach LLMs to learn translation. However, all these methods neglect the importance of instruction finetuning data quality and diversity in machine translation. And in this paper, we propose a novel approach to enhance the quality and diversity of translation data.

\paragraph{Traning Data Quality and Diversity.} Various studies evident that the quality and diversity of instruction finetuning data predominate the performance of LLMs \citep{zhou2023lima,touvron2023llama,li2023quantity}. For example, \citet{zhou2023lima} manually curated a small, high-quality instruction set to elevate the model’s instruction following power. Although the methods in \citep{zhou2023lima} rely heavily on human effort, they motivate research aiming to automatically obtain high quality instructions. \citet{cao2023instruction} propose to score the quality of each instruction by combining several language indicators using a linear model. However, all these indicators rely on external models. Besides, \citet{du2023mods} present a comprehensive approach for selecting high quality and diverse instruction based on reward model score and semantic diversity. Still, their methods rely heavily on external models and overlook the direct impact that finetuning data has on the model. Remarkably, \citet{li2023quantity} propose a self-guided approach to select difficult instructions with the guidance of the LLM itself. Admittedly, their methods are free of any external models. However, they select training examples more associated with necessity and complexity than quality, and they overlook the importance of finetuning data diversity. Despite the efforts these methods have made to automate instruction selection, they either overlook the importance of the quality and diversity of training data or rely on external models for judging. Significantly different from them, our methods take both quality and diversity into consideration and are free of external models.
\begin{figure*}[!th]
    \centering
    \includegraphics[width=\textwidth]{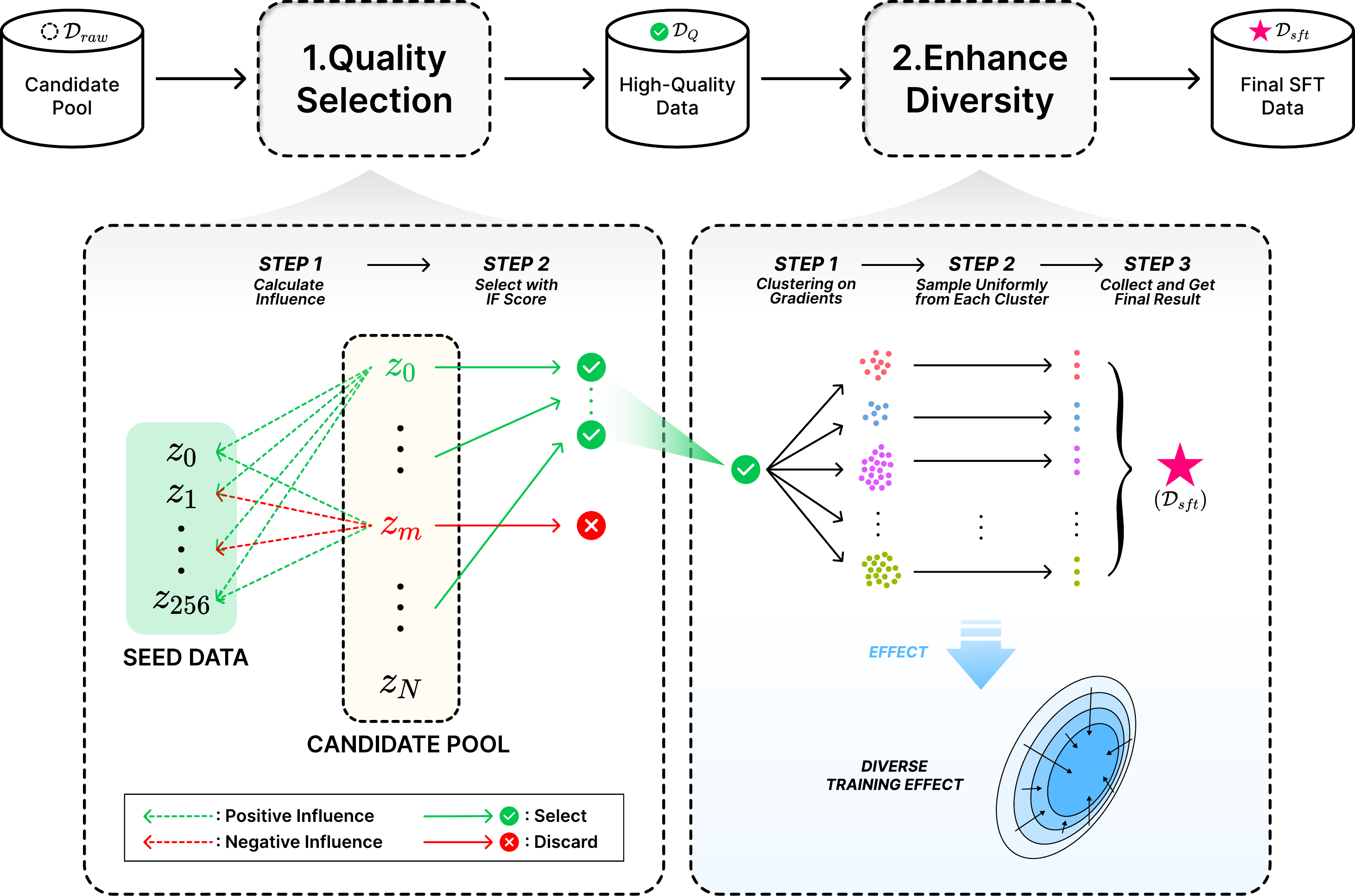}
    \caption{Overview of our proposed method. Our overall method consists of two components: (1) high-quality data selection and (2) enhancing their diversity. In high-quality data selection, we calculate the pair-wise influence (dashed arrows) of the candidates on seed data. Then we select those with all positive influences (as marked green). Afterwards, we cluster on the selected high-quality data to distinguish dissimilar influences (as marked in dots with different colors)
    and resample to further obtain high-quality and diverse finetuning data.}
    \label{fig:overview}
\end{figure*}
\paragraph{Gradient-based Data Selection.} Data selection with influence function has been widely studied in NLP.~\citet{lam-etal-2022-analyzing} propose to identify erroneous training data by using synthetic noisy data, showing that vanilla influence functions are not sufficient for good retrieval performance. On the contrary, we select high-quality and diverse finetuning data with the aid of gradient information. Also, we show that our use of influence function is capable of selecting high-quality data.~\citet{akyurek-etal-2022-towards} demonstrate the potential of using gradient information to trace factual knowledge in language models back to the training data. Nevertheless, their practical applications remain understudied. Methods for scaling up influence function \cite{schioppa2022scaling} and explaining black box predictions of NLP models have been proposed~\cite{han-etal-2020-explaining}. Remarkably, the concurrent work \cite{xia2024less} proposes a similar method to select task-specific LLM finetuning data by estimating influences for training data. Different from them, we select high-quality finetuning data for machine translation. And we show that our method is capable of capturing a higher-level concept (i.e., the quality of training data).

\section{Methods}
\label{sec:Methods}
In this section, we describe our gradient-based method to select high-quality and diverse training data for instruction finetuning, as displayed in Figure~\ref{fig:overview}. Our methods consist of two parts: 1) high-quality data selection with the influence function (\S~\ref{sec:qual}) and 2) diverse data selection with gradient-based clustering (\S~\ref{sec:dive}). For high-quality data selection, importantly, we utilize the influence function to quantify the impact of individual training on the test sample. For diverse data selection, we use gradient distance to assess the overall diversity of the instruction training data.

\subsection{High-quality Data Selection}
\label{sec:qual}
In this section, we detail our approach for selecting high-quality training data for machine translation. Intuitively, if a training example significantly benefits the model to generate high-quality outputs, it is likely to be of high quality itself. Consequently, we first manually curate a small set of high-quality translation data that we refer to as the \emph{seed data} to form a criteria for evaluating training data quality. Then we select training data that aids the model in generating high-quality seed data. 

Concretely, we employ Influence Function (IF) \citep{koh2017understanding} to quantify how a training example $z_m$ influences the model's behavior on a test example $z_t$. In our influence function setting, we start with the following finetuning objective:
\begin{equation}
    \label{equ:finetuning-obj}
\bm{\theta}^*:
    =\mathop{\arg\min}_{\bm{\theta}}\frac{1}{n}\sum_{i=1}^nL(z_i|\bm{\theta}),
\end{equation}
where $\bm\theta$ is the model parameter, $z_i=(x_i,y_i)$ is the $i$-th prompt-response pair and the loss $L$ is simply the language modeling loss of the response solely:
\begin{align}
\label{equ:loss}
    L(x, y)&=-\log P(y|x) \notag \\
    &=-\log\prod_{j=1}^{T}p(y^{j}|x, y^{<j}),
\end{align}
where $x^j$ denotes the $j$-th token of $x$. Then influence function calculates the influence of $z_m$ on $z_t$ by:
\begin{equation}
    \mathcal{I}(z_m, z_t)=-\nabla_{\bm{\theta}|\bm{\theta}^*}L(z_t)^\top{\bf H}^{-1}_{\bm{\theta}^*}\nabla_{\bm{\theta}|\bm{\theta}^*}L(z_m),
\label{equ:if_cf}
\end{equation}
where $\bm{\theta}^*$ is the optimal model parameter of finetuning in \eqref{equ:finetuning-obj} and ${\bf H}_{\bm{\theta}^*}=\nabla_{\bm{\theta}}^2\frac{1}{n}\sum_{i=1}^n L(z_i|\bm{\theta}^*)$ is the Hessian of the training objective at $\bm{\theta}=\bm{\theta}^*$. 
Though the calculation of the influence function is complex, the most significant aspect for readers is that \emph{if the scalar $\mathcal{I}(z_m, z_t)$ is negative, training the model on $z_m$ reduces the model's loss on $z_t$. In this case, $z_m$ is considered to be helpful for the model to generate $z_t$}. In our implementation of IF, we use the gradients of model's Multilayer Perceptron (MLP) parameters in $\{3,6,9,12,15,18,21,24,27,30\}$-th layers to speed up calculation. And we average the gradients of each token to form a vector. Also, we use Kronecker-Factored Approximate Curvature (KFAC) \citep{martens2015optimizing} to approximate Hessian for reducing memory consumption. We detail the derivation of IF and our modification for fitting it into LLM fining tuning in Appendix~\ref{app:if}.


Correspondingly, as depicted in Figure \ref{fig:overview}, our proposed high-quality data selection method consists of two steps: (1) calculating influences and (2) selection.
We select training examples in our candidate pool $\mathcal{D}_{raw}$ that exert beneficial impact on all samples in the seed data, i.e., example $z_m\in\mathcal{D}_{raw}$ meets:
\begin{equation}
    \forall z_t\in\mathcal{D}_{seed}, ~\mathcal{I}(z_m, z_t)<0.
    \label{equ:critiea}
\end{equation}
Thus, we select training data that contributes to the model's high-quality generation.

In practice, we find that the seed instruction dataset $\mathcal{D}_{seed}$ of size 256 suffices for selecting high quality instructions. Hence, we set the size of $\mathcal{D}_{seed}$ to 256 in our implementation. As the focus of this paper is to select high-quality finetuning data for translation task, we randomly select 256 parallel texts from the validation set and have them revised by human translators. We select high-quality data from our candidate pool, for which we collect publicly available WMT22's parallel texts. Throughout our implementations, we use the prompt template: \textit{Translate the following text into \{trg\_lang\}.\textbackslash n\textbackslash nText:\textbackslash n``\{src\_text\}''}, where \textit{\{trg\_lang\}} are target languages such as "English" and \textit{\{src\_text\}} are the source text. And the response is simply the target text. We detail our data preparation in Section~\ref{sec:exp-setup}.


\subsection{Diverse Data Selection}
\label{sec:dive}
After obtaining high-quality translation pairs, we further ensure the diversity of the training data. To ensure coverage of different translation patterns, we propose to use gradient similarity to assess the diversity. Specifically, we consider the gradient of the response loss in equation \eqref{equ:loss} with respect to the final MLP layer and average them out on all the tokens. We utilize the Euclidean distance as the similarity measure.

To maximize the diversity of training data influences, we cluster on the gradients of training examples to obtain different patterns. Then we sample uniformly from the clustering result to ensure the diversity of training data.
Moreover, we employ \emph{K-means} as the clustering algorithm due to its ability to process large-scale datasets. Furthermore, to speed up and reduce memory, we use \emph{Random Projection} as the dimensionality reduction method to reduce the dimension of the gradients to 400~\cite{bingham2001random}. In practice, we cluster the training data into 512 clusters.

\section{Experiment Setup}
\label{sec:exp-setup}
\begin{figure*}
    \centering
    \includegraphics[width=\linewidth]{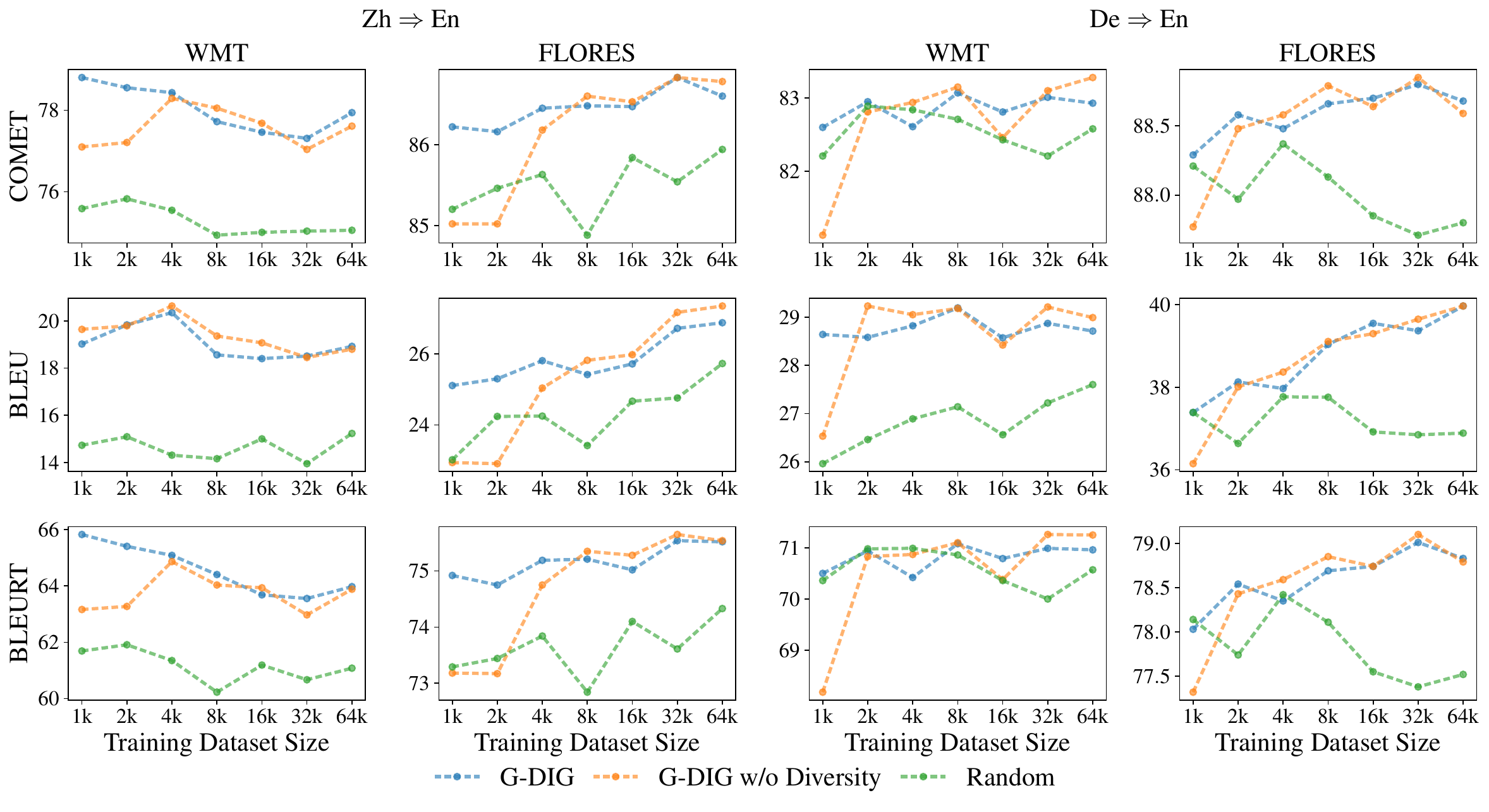}
    \caption{The comparison results of model trained on various amounts of data selected by our \method, \methodq and Random selection on Zh $\Rightarrow$ En and De $\Rightarrow$ En translations. We plot the results on Zh $\Rightarrow$ En and De $\Rightarrow$ En translations in the left and right two columns respectively.}
    \label{fig:maingraph}
\end{figure*}
\paragraph{Datasets.}
We conduct experiments on Zh $\Rightarrow$ En and De $\Rightarrow$ En tasks. We collect separate candidate pools for different translation directions. Specifically, our Zh $\Rightarrow$ En pool is composed of 1.9 million examples sampled from WMT22's ParaCrawl v9, News Commentary v16, UN Parallel Corpus V1.0, WikiMatrix and Back-translated news and our De $\Rightarrow$ En pool contains 256k examples sampled from WMT22's Europarl v10, ParaCrawl v9, News Commentary v16 and WikiMatrix. We split 512 examples to validation sets for evaluation. We test our methods on the latest WMT22 test sets from the news translation track of WMT22 competition\footnote{https://github.com/wmt-conference/wmt22-news-systems} and Flores-101 dev-test split \citep{goyal2022flores}. The WMT22 Zh $\Rightarrow$ En and De $\Rightarrow$ En test sets contain 1875 and 1984 samples, respectively. And the Flores-101 dev-test split contains 1012 samples for each Zh $\Rightarrow$ En and De $\Rightarrow$ En.


\paragraph{Implementation Details.}
 We use Baichuan2-7B \citep{baichuan2023baichuan2} for Zh $\Rightarrow$ En and Llama2-7B for De $\Rightarrow$ En. 
 For all finetuning experiments, we adopt the same setting.  The finetuning process lasts for 3 epochs with an initial learning rate of $1e-5$ and a global batch size of 64. 
 The instruction template we use is \textit{Translate the following text into \{trg\_lang\}.\textbackslash n\textbackslash nText:\textbackslash n``\{src\_text\}''}, where \textit{\{trg\_lang\}} are target languages such as "English" and \textit{\{src\_text\}} are the source text.
The models are evaluated every 10 steps. We use the checkpoint with the smallest loss on the valid set for the final test. During the inference, we use beam search as the decoding strategy with a beam size of 4. The training and inference of 7B size models are conducted on 16 NVIDIA A100 80GB GPUs with DeepSpeed ZeRO-3 Offload.



\paragraph{Evaluation.}

For automatic evaluation, we use the widely used metrics BLEU \citep{papineni2002bleu}, BLEURT \citep{sellam2020bleurt}, and COMET \citep{rei2020comet}. We use ScareBLEU\footnote{https://github.com/mjpost/sacrebleu}, BLEURT-20 \citep{pu2021learning} and Unbabel/wmt22-comet-da\footnote{https://huggingface.co/Unbabel/wmt22-comet-da}  in the evaluation implementations.

\begin{table*}[!ht]\Huge
\renewcommand    
	\arraystretch{1.3}
\resizebox{\linewidth}{!}{
\begin{tabular}{l c c c c c c c c c c c c}
\toprule
\multirow{3}{*}{Model} & \multicolumn{6}{c}{Zh $\Rightarrow$ En} & \multicolumn{6}{c}{De $\Rightarrow$ En}       
\\ \cmidrule(r){2-7} \cmidrule(r){8-13} 
                       & \multicolumn{3}{c}{WMT22}                  & \multicolumn{3}{c}{FLORES}                 & \multicolumn{3}{c}{WMT22} & \multicolumn{3}{c}{FLORES} \\ \cmidrule(r){2-4} \cmidrule(r){5-7} \cmidrule(r){8-10} \cmidrule(r){11-13} 
                       & {COMET} & BLEU  & BLEURT & {COMET} & BLEU  & BLEURT & COMET  & BLEU   & BLEURT  & COMET   & BLEU   & BLEURT  \\ \midrule
\textit{SOTA Models}   &                           &       &        &                           &       &        &        &        &         &         &        &         \\
Bayling-13B            & 77.72                     & 20.12 & -      & -                         & -     & -      & \textbf{83.02}  & \textbf{27.34}  & -       & -       & -      & -       \\
BigTranslate-13B       & 74.26                     & 14.16 & -      & -                         & -     & -      & 80.68  & 23.35  & -       & -       & -      & -       \\
TIM-7B                 & \colorbox{pink}{\textbf{79.33}}                     & \colorbox{pink}{\textbf{23.81}} & -      & \textbf{85.81}                     & \colorbox{pink}{\textbf{26.25}} &        & 78.19  & 25.43  & -       & \colorbox{pink}{\textbf{88.82}}   & \colorbox{pink}{\textbf{41.96}}  & -       \\
NLLB-54B               & 70.70                     & 16.56 & -      & -                         & -     & -      & 78.94  & 26.89  & -       & -       & -      & -       \\ \midrule
\textit{Baseline Models} &                           &       &        &                           &       &        &        &        &         &         &        &         \\
Random              & 75.55                     & 14.31 & 61.35  & 85.63                     & 24.25 & 73.84  & 82.58  & 27.60  & 70.57   & 87.80   & 36.89  & 77.52   \\
Reward              & 77.90                     & 16.29 & 64.64  & 86.16                     & 24.48 & 74.73  & \colorbox{pink}{\textbf{83.28}}  & 27.52  &      \colorbox{pink}{\textbf{71.51}}   & 88.37   & 38.69  & 78.62   \\ 
\textit{Ours}          &                           &       &        &                           &       &        &        &        &         &         &        &         \\
\method             & 78.29                     & \textbf{20.63} & 64.86  & 86.18                     & 25.04 & 74.75  & \colorbox{pink}{\textbf{83.28}}  & \colorbox{pink}{\textbf{28.99}}  & 71.25   & 88.59   & \textbf{39.97}  & 78.79   \\
\methodq   & \textbf{78.43}                     & 20.35 & \colorbox{pink}{\textbf{65.08}}  & \colorbox{pink}{\textbf{86.45}}                     & \textbf{25.81} & \colorbox{pink}{\textbf{75.19}}  & 82.93  & 28.71  &      70.96   & \textbf{88.68}   & \textbf{39.97}  & \colorbox{pink}{\textbf{78.83}}   \\ \bottomrule
\end{tabular}
}
\caption{In this table, we present the comparison results of our methods with various baselines in accordance with Section \ref{sec:exp-setup}. We directly adopt the results from the original paper and omit the missing metrics. We report the results of our \method and \methodq. The Best result in each group is in \textbf{bold}. The Best result in each column is in \colorbox{pink}{{red}}.}
\label{tb:main-tb}

\end{table*}

\paragraph{Baselines and Comparisons.} In order to demonstrate the superiority and effectiveness of our methods, we compare our \method to validate the performance of our overall approach. Also, we assess its variant \methodq (\method without enhancing the data diversity) to solely investigate our high-quality data selection module. For comparisons, we consider the following baselines:
\begin{itemize}
    \item Bayling-13B \citep{bayling}, an English / Chinese LLM based on Llama with superior translation capabilities.
    \item BigTranslate-13B \citep{yang2023bigtrans}, a multilingual LLM based on Llama with the capability of translating over 100 natural languages.
    \item TIM \citep{zeng2023tim}. We present the results of BLOOMZ-7b-mt and Llama2-7b trained with TIM-Full-LM-based Bad Output for WMT22 test sets and FLORES respectively. 
    \item Model trained on the random subset. To emphasize the need for instruction selection, we choose $k$ random instructions to form a finetuning subset. We finetune the LLM on the selected subsets and report their performances.
    \item Model trained on the reward subset. We also compare our method with the existing selection method. We use the commonly used reward model-based method for selecting high-quality training data \citep{du2023mods,cao2023instruction}. Specifically, we follow \citep{du2023mods} to use the {reward-model-debertav3-large-v2}\footnote{https://huggingface.co/OpenAssistant/reward-model-deberta-v3-large-v2} to score each instruction and select the top $k$ instructions with the highest score to form the finetuning subset.
\end{itemize}

\section{Experimental Results}

\subsection{Main Results}
In this section, we present our main experimental findings. We start with comparing our methods with baselines on various amounts of training data in Figure \ref{fig:maingraph}. Then we compare our best results with SOTA models and baselines in Table \ref{tb:main-tb}. Furthermore, we conduct human evaluation and present the results in Table \ref{Table_Results_Human}. We show that our approach is superior in terms of effectiveness and robustness.

\paragraph{Our Methods Improve the Translation Performance Across Various Amount of Training Data.}
In order to demonstrate the scalability of our method in terms of the amount of selected training data, we present the results as a function of the amount of training data from 1000 to 64,000 in Figure \ref{fig:maingraph} for Zh $\Rightarrow$ En and De $\Rightarrow$ En translation. Notably, our \method model consistently surpasses the random model across \emph{all} metrics and dataset sizes for Zh $\Rightarrow$ En translation. Also, for De $\Rightarrow$ En translation, our \method outperforms the random baseline in almost all cases. Statistical analysis results in Appendix \ref{app:stat} further suggest our methods excel the random baseline. These results demonstrate the efficacy and robustness of our methods in terms of the amount of selected data. Meanwhile, we also observe that the quality and diversity of finetuning data dominate the performance of LLM. Specifically, for both Zh $\Rightarrow$ En and De $\Rightarrow$ En translations we can see that models trained on 1000 examples selected by our \method outperform the models trained with 64k randomly selected examples.
\begin{figure*}[!t]
    \centering
    \includegraphics[width=\linewidth]{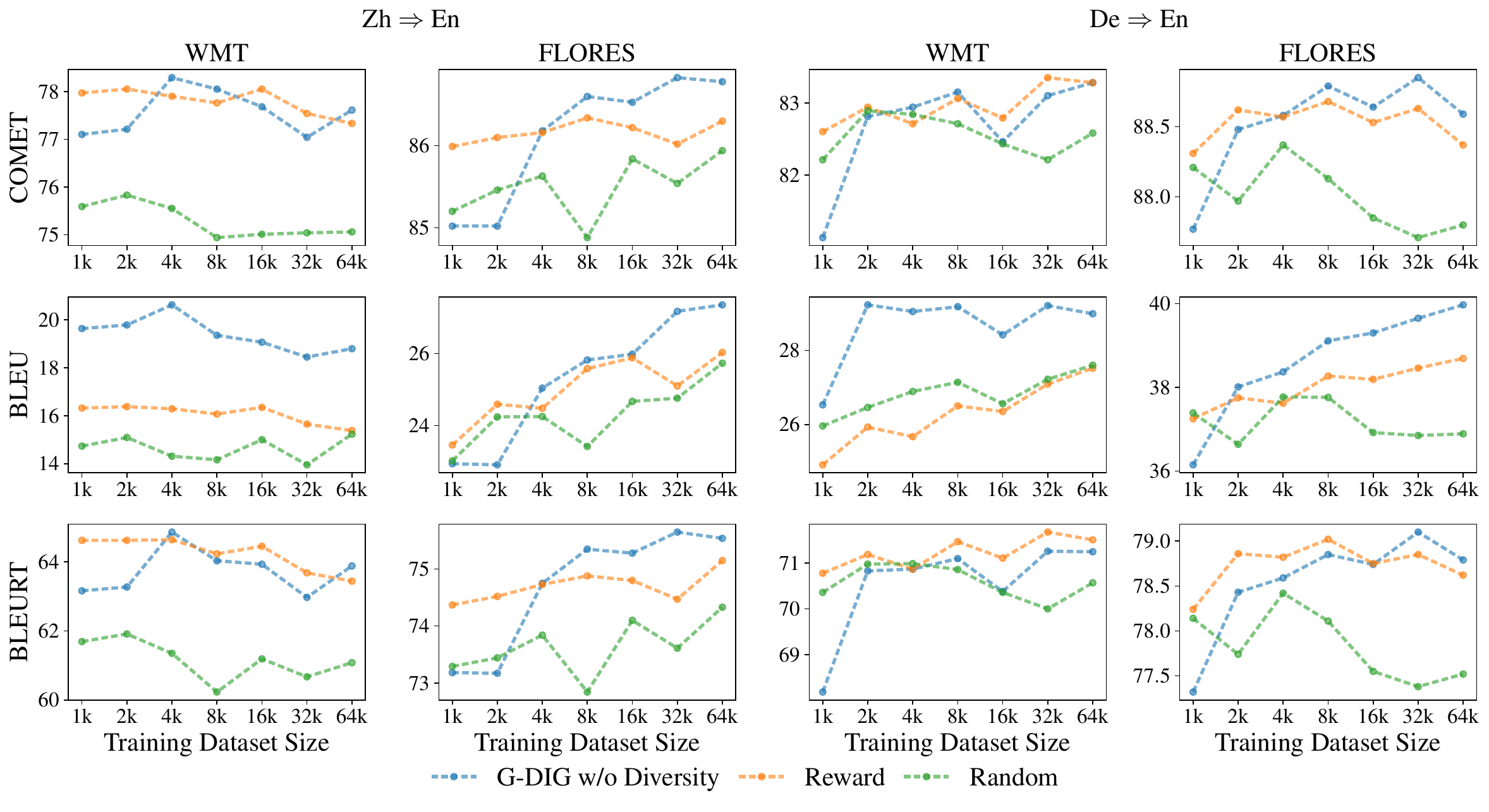}
    \caption{The comparison results of model trained on various amounts of data selected by our \methodq, Reward model selection and Random selection on Zh $\Rightarrow$ En and De $\Rightarrow$ En translations. We plot the results on Zh $\Rightarrow$ En and De $\Rightarrow$ En translations in the left and right two columns respectively.}
    \label{fig:ab-quality}
\end{figure*}
\begin{table}[t]\small
	\renewcommand    
	\arraystretch{1.3}
	\centering
        \resizebox{\columnwidth}{!}{
            \setlength{\tabcolsep}{1.5mm}
    	\begin{tabular}{ l  c c c c}
    		\toprule
    		  Model & Score & Win $\uparrow$ & Tie & Lose $\downarrow$ \\
            \midrule
      \multicolumn{5}{c}{Zh $\Rightarrow$ En} \\
        \midrule
              w/ Random & 3.59 &   --- & --- &  ---  \\
              w/ Ours & 3.92 &  35.0\% & 53.0\%  &  12.0\%  \\
              \midrule
              \multicolumn{5}{c}{De $\Rightarrow$ En} \\
        \midrule
              w/ Random & 3.92 &   --- & --- &  ---  \\
              w/ Ours & 4.21 &  34.0\% & 53.0\%  &  13.0\%  \\
            
    	\bottomrule
    	\end{tabular}
     }
	\caption{
		\label{Table_Results_Human}
 	Human evaluation results on randomly sampled sets.  ``Win''/``Tie''/``Lose'' stands for the percentage of translations where ours is better than, tied with, or worse than the random subset.}
\end{table}


\paragraph{Our Methods Surpass Baselines and Achieve Comparable Results with SOTA Models.}
To see how our method performs compared with baselines and SOTA models, in Table \ref{tb:main-tb} we present the results of SOTA models, our best and corresponding baselines. Specifically, our \method achieves its best results at 4000 training examples for Zh $\Rightarrow$ En translation and 64k for De $\Rightarrow$ En translation. For baselines comparisons, we observe that our methods surpass baselines in almost all cases, demonstrating the effectiveness of our approach. For SOTA models comparisons, our 7B models achieve comparable results with TIM-7B and even better results compared with Bayling-13B, BigTranslate-13B and NLLB-54B.

\paragraph{Our Methods Align the Model Better Compared with Random Baseline.}
We further conduct human evaluation to analyze the translation quality. We respectively randomly pick 100 sentences from Zh $\Rightarrow$ En and De $\Rightarrow$ En test sets, and recruit three human judges for evaluation. For each sentence, the judges read the source sentence and two candidate translations, which are from the random subset model and \method subset model. The judges are required to rate each candidate on a scale of 1 to 5 and pick the better one.

From Table~\ref{Table_Results_Human}, we can see our methods enable the model to translate better with a higher average score in both Zh $\Rightarrow$ En and De $\Rightarrow$ En translations. Also, our \method subset model is more frequently rated as better translation than the random subset model on both Zh $\Rightarrow$ En and De $\Rightarrow$ En translations, indicating our methods better align the model than the random selection method.

\paragraph{The Fewer the Instructions, the Greater the Significance of Diversity.}
To see the role that diversity plays during finetuning, we compare the results of \method and \methodq on Zh $\Rightarrow$ En and De $\Rightarrow$ En translation in Figure \ref{fig:maingraph}. Remarkably, our diversity enhancement makes significant advancements in enhancing the translation performance when there are only few training data provided ($k\le2000$). For Zh $\Rightarrow$ En translation with 1000 training examples, our \method further improves \methodq by 1.7 in terms of COMET on WMT and by 2.17 in terms of BLEU on FLORES. In addition, for De $\Rightarrow$ En translation with 1000 training examples, our \method improves the BLEU by 2.11 and 1.24 on WMT and FLORES respectively compared with \methodq. However, as the amount of instructions increases, this effect fades away, since large instruction sets are already coupled with high diversity. As shown in Figure \ref{fig:maingraph}, the \method curves almost coincide with the \methodq curves in all metrics when the amount of instructions goes beyond 4000 and 8000 for WMT and FLORES, respectively.




\subsection{Analysis}
\begin{table*}[!t] \small
    \centering
    \resizebox{\linewidth}{!}{
    \begin{CJK*}{UTF8}{gkai}
    \begin{tabular}{ p{0.25\linewidth} p{0.25\linewidth} p{0.25\linewidth} p{0.25\linewidth}}
    
    \toprule

    \multicolumn{2}{c}{Zh $\Rightarrow$ En} & \multicolumn{2}{c}{De $\Rightarrow$ En} \\
    \cmidrule(r){1-2}  \cmidrule(r){3-4} 
          Source Text & Target Text &Source Text & Target Text \\
        \hline \textit{Ours}
        \\
        例如，如果在大部分时间里价格在1,200美元到1,800美元之间，这就是常见值域。
         &For example, if the price is between \$1,200 and \$1,800 most of the time, this is the common range. &Der ganze MDF Italia Tense Tisch wird mit einer 3mm dünnen Acrylharzfolie und Steinmineralien in Weiß verkleidet. & The entire MDF Italia Tense Table is coated with a 3 mm thin acrylic resin film and stone minerals in white. \\ 
         \\
         根据武器法第6条，若无司法部长颁发许可证，禁止出口(任何种类)武器和武器设备。 & According to the Weapons Act, Section 6, it is prohibited without a license from the Minister of Justice to export weapons (of any kind) and war equipment. & X850 Infrarot-Induktions-Funk-Türklingel w / blaue Anzeige - Weiß (3 x AAA + 3 x AA) & X850 Infrared Induction Wireless Doorbell w/ Blue Indicator - White (3 x AAA + 3 x AA) \\ \hline \textit{Random} \\
         在对她的行为进行调查之前，西拉德福德议员被剥夺了议会的鞭打并被禁止参加政党活动。&The West Bradford MP was stripped of the parliamentary whip and barred from party activity pending an investigation of her behaviour - which David Cameron branded racist.&Ebenso wichtig ist das Engagement in interdisziplinären und internationalen Forschungszweigen wie der Critical Psychology, den Cultural und Postcolonial Studies, den Gender wie auch den Religious Studies.& In this regard, sociology, anthropology, history, education, and philosophy are considered as important as work in research fields like critical psychology, cultural and postcolonial studies, gender studies, and religious studies.\\
        \\
         我们提供2134 个酒店在西安 & Hotels in Xi'an (Shaanxi) &Hin- und Rückflüge von Lagos nach Madrid&Return flights from Lagos to Madrid \\
         \bottomrule
        
    \end{tabular}
    \end{CJK*}
    }
    \caption{Examples of ours and randomly selected training data in the form of parallel texts.}
    \label{tab:case-quality}
\end{table*}

\begin{figure}[t!]
    \centering
    \includegraphics[width=\linewidth]{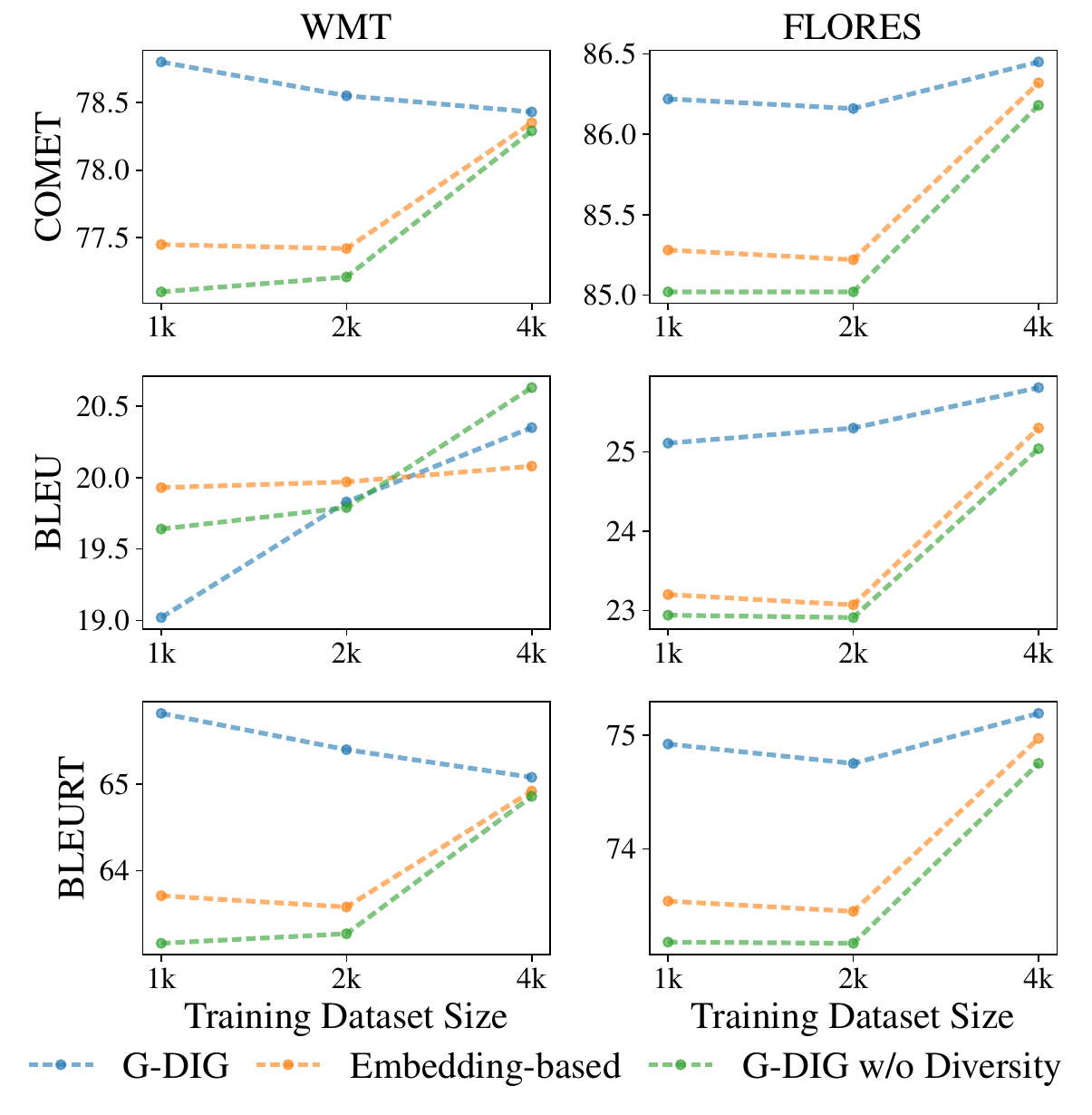}
    \caption{The comparisons between our \method,\methodq and embedding-based methods on various amounts of training data on Zh $\Rightarrow$ En translation. 
    }
    \label{fig:ab-grad}
\end{figure}

In this section, we conduct ablation studies on our high-quality data selection module and our diversity enhancement module. Specifically, to see the superiority of our high-quality data selection module, we compare  \methodq with the reward model-based baseline in Figure \ref{fig:ab-quality}. In order to demonstrate the advantage of our gradient-based diverse data selection module, we compare our \method with its embedding-based counterparts in Figure \ref{fig:ab-grad}. 
\paragraph{Our High-quality Data Selection Module Surpasses the Reward Model-based Method}
In Figure \ref{fig:ab-quality}, we show that our high-quality data selection module \methodq achieves superior results compared with the existing reward model-based method in Figure \ref{fig:ab-quality}. Specifically, for Zh $\Rightarrow$ En translation our \method achieves overall better results compared with the reward model-based method, falling behind in only few cases. For De $\Rightarrow$ En translation, it is hard to distinguish between ours and reward models-based baselines in COMET and BLEURT. However, our \methodq significantly outperforms the reward model in terms of BLEU on both WMT22 and FLORES.

\paragraph{Our Diversity Enhancement Module Improves the Training Data Diversity.}

In this part, we substantiate the superiority of our gradient-based diversity enhancement module. We compare our \method with its embedding-based counterpart. Specifically, embedding-based methods measure the similarity of training examples based on embeddings. Their objective is to maximize the semantic richness of the training data. We follow the model used in \citep{du2023mods} for diversifying the training data. We use BERT \citep{devlin2018bert} to extract embeddings. Then we run \emph{Kmeans} to cluster these embeddings and use our sampling procedure to enhance the training data diversity. As illustrated in Figure~\ref{fig:ab-grad}, our proposed method outperforms its embedding-based counterpart in almost all cases, which validates the advantage of our gradient-based method. For instance, with 1000 instructions provided, our methods surpass the embedding-based method by 2.11 of BLEURT and 1.91 of BLEU on WMT and FLORES respectively. Moreover, in almost all scenarios, the curves of embedding-based method coincide with \methodq curves with trivial improvements, indicating that the existing methods that enhance semantic coverage contribute little to the finetuning data diversity.

\paragraph{Our Methods Select Highly Parallel Texts.}
In this part, we showcase our selected high-quality data and randomly selected data in the form of parallel texts in Table \ref{tab:case-quality}. Remarkably, not only is our selected data accurate and coherent in the target text space, but it is also natural and of the correct format and grammar in terms of the source text.

\subsection{Discussion on Hyperparameters}
There are two primary hyperparameters in our methods: the number of seed data $\vert\mathcal{D}_{seed}\vert$ in high-quality data selection and the number of clusters for K-means in diverse data selection. Intuitively, the more seed data we use, the more strict the criteria in \eqref{equ:critiea} would be. Therefore, increasing the number of seed data improves the quality of selected data. And using less seed data can be regarded as a relaxation to the criteria in~\eqref{equ:critiea}. We experiment with seed dataset sizes of 128, 256, and 512 and note that further enlarging the seed dataset contributes little to improving the quality of selected seed data. Considering the high cost of obtaining human-annotated high-quality data, we ultimately use 256 seed data after balancing annotation costs and model performance benefits.

The reasonable choice of the number of clusters in diverse data selection is crucial for our methods. Since our candidate pool is quite large, we cluster the data into 512 clusters to ensure the variety of training data.

\section{Conclusion}
In this paper, we propose \method, a novel gradient-based method for selecting high-quality and diverse LLM finetuning data for machine translation. Specifically, using Influence Function and a seed dataset we select high-quality training data that have beneficial influence on the model. Furthermore, we enhance the training data diversity by clustering on their gradients and resampling. Extensive experiments prove that our methods improve the LLM in terms of translation ability. Also, human evaluation results demonstrate that our methods better align the LLM compared with the baseline model. We hope this work facilitates the research on LLM finetuning data selection. 

\section*{Limitations}

\paragraph{Computational Cost for Influence Function.}
In this paper, we utilize Influence Function to measure the influence of a training sample on a test sample's prediction. However, the computational cost for calculating this influence can be large for LLMs. The asymptotic computational complexity for calculating Hessian is $\mathcal{O}(P^2)$ where $P$ is the number of model parameters. And the computational complexity for calculating pair-wise influence between the seed data and candidate pool is $\mathcal{O}(MNP)$ where $M$ and $N$ denote the number of seed data and candidates, respectively. How to reduce the computational cost of influence function still remains a challenge. We leave this for future work. 

\section*{Ethical Considerations}
All the data sources are publicly available and do not involve privacy issues. For all human evaluations mentioned in the paper, we hire full-time professional translators and pay them with market wage. All of our translators are graduates.

\bibliography{main}

\begin{thebibliography}{45}
\expandafter\ifx\csname natexlab\endcsname\relax\def\natexlab#1{#1}\fi

\bibitem[{Achiam et~al.(2023)Achiam, Adler, Agarwal, Ahmad, Akkaya, Aleman, Almeida, Altenschmidt, Altman, Anadkat et~al.}]{achiam2023gpt}
Josh Achiam, Steven Adler, Sandhini Agarwal, Lama Ahmad, Ilge Akkaya, Florencia~Leoni Aleman, Diogo Almeida, Janko Altenschmidt, Sam Altman, Shyamal Anadkat, et~al. 2023.
\newblock Gpt-4 technical report.
\newblock \emph{arXiv preprint arXiv:2303.08774}.

\bibitem[{Agrawal et~al.(2022)Agrawal, Zhou, Lewis, Zettlemoyer, and Ghazvininejad}]{agrawal2022context}
Sweta Agrawal, Chunting Zhou, Mike Lewis, Luke Zettlemoyer, and Marjan Ghazvininejad. 2022.
\newblock In-context examples selection for machine translation.
\newblock \emph{arXiv preprint arXiv:2212.02437}.

\bibitem[{Akyurek et~al.(2022)Akyurek, Bolukbasi, Liu, Xiong, Tenney, Andreas, and Guu}]{akyurek-etal-2022-towards}
Ekin Akyurek, Tolga Bolukbasi, Frederick Liu, Binbin Xiong, Ian Tenney, Jacob Andreas, and Kelvin Guu. 2022.
\newblock Towards tracing knowledge in language models back to the training data.
\newblock In \emph{Findings of the Association for Computational Linguistics: EMNLP 2022}, pages 2429--2446, Abu Dhabi, United Arab Emirates. Association for Computational Linguistics.

\bibitem[{Baichuan(2023)}]{baichuan2023baichuan2}
Baichuan. 2023.
\newblock Baichuan 2: Open large-scale language models.
\newblock \emph{arXiv preprint arXiv:2309.10305}.

\bibitem[{Bingham and Mannila(2001)}]{bingham2001random}
Ella Bingham and Heikki Mannila. 2001.
\newblock Random projection in dimensionality reduction: applications to image and text data.
\newblock In \emph{Proceedings of the seventh ACM SIGKDD international conference on Knowledge discovery and data mining}, pages 245--250.

\bibitem[{Cao et~al.(2023)Cao, Kang, Wang, and Sun}]{cao2023instruction}
Yihan Cao, Yanbin Kang, Chi Wang, and Lichao Sun. 2023.
\newblock Instruction mining: When data mining meets large language model finetuning.
\newblock \emph{arXiv preprint arXiv:2311.15653}.

\bibitem[{Chen et~al.(2023{\natexlab{a}})Chen, Li, Yan, Wang, Gunaratna, Yadav, Tang, Srinivasan, Zhou, Huang, and Jin}]{chen2023alpagasus}
Lichang Chen, Shiyang Li, Jun Yan, Hai Wang, Kalpa Gunaratna, Vikas Yadav, Zheng Tang, Vijay Srinivasan, Tianyi Zhou, Heng Huang, and Hongxia Jin. 2023{\natexlab{a}}.
\newblock \href {http://arxiv.org/abs/2307.08701} {Alpagasus: Training a better alpaca with fewer data}.

\bibitem[{Chen et~al.(2023{\natexlab{b}})Chen, Liu, Meng, Chen, Xu, and Zhou}]{chen2023improving}
Yijie Chen, Yijin Liu, Fandong Meng, Yufeng Chen, Jinan Xu, and Jie Zhou. 2023{\natexlab{b}}.
\newblock Improving translation faithfulness of large language models via augmenting instructions.
\newblock \emph{arXiv preprint arXiv:2308.12674}.

\bibitem[{Chung et~al.(2022)Chung, Hou, Longpre, Zoph, Tay, Fedus, Li, Wang, Dehghani, Brahma et~al.}]{chung2022scaling}
Hyung~Won Chung, Le~Hou, Shayne Longpre, Barret Zoph, Yi~Tay, William Fedus, Yunxuan Li, Xuezhi Wang, Mostafa Dehghani, Siddhartha Brahma, et~al. 2022.
\newblock Scaling instruction-finetuned language models.
\newblock \emph{arXiv preprint arXiv:2210.11416}.

\bibitem[{Devlin et~al.(2018)Devlin, Chang, Lee, and Toutanova}]{devlin2018bert}
Jacob Devlin, Ming-Wei Chang, Kenton Lee, and Kristina Toutanova. 2018.
\newblock Bert: Pre-training of deep bidirectional transformers for language understanding.
\newblock \emph{arXiv preprint arXiv:1810.04805}.

\bibitem[{Du et~al.(2023)Du, Zong, and Zhang}]{du2023mods}
Qianlong Du, Chengqing Zong, and Jiajun Zhang. 2023.
\newblock Mods: Model-oriented data selection for instruction tuning.
\newblock \emph{arXiv preprint arXiv:2311.15653}.

\bibitem[{Garcia et~al.(2023)Garcia, Bansal, Cherry, Foster, Krikun, Johnson, and Firat}]{garcia2023unreasonable}
Xavier Garcia, Yamini Bansal, Colin Cherry, George Foster, Maxim Krikun, Melvin Johnson, and Orhan Firat. 2023.
\newblock The unreasonable effectiveness of few-shot learning for machine translation.
\newblock In \emph{International Conference on Machine Learning}, pages 10867--10878. PMLR.

\bibitem[{Goodfellow et~al.(2016)Goodfellow, Bengio, and Courville}]{goodfellow2016deep}
Ian Goodfellow, Yoshua Bengio, and Aaron Courville. 2016.
\newblock \emph{Deep learning}.
\newblock MIT press.

\bibitem[{Goyal et~al.(2022)Goyal, Gao, Chaudhary, Chen, Wenzek, Ju, Krishnan, Ranzato, Guzm{\'a}n, and Fan}]{goyal2022flores}
Naman Goyal, Cynthia Gao, Vishrav Chaudhary, Peng-Jen Chen, Guillaume Wenzek, Da~Ju, Sanjana Krishnan, Marc’Aurelio Ranzato, Francisco Guzm{\'a}n, and Angela Fan. 2022.
\newblock The flores-101 evaluation benchmark for low-resource and multilingual machine translation.
\newblock \emph{Transactions of the Association for Computational Linguistics}, 10:522--538.

\bibitem[{Han et~al.(2020)Han, Wallace, and Tsvetkov}]{han-etal-2020-explaining}
Xiaochuang Han, Byron~C. Wallace, and Yulia Tsvetkov. 2020.
\newblock Explaining black box predictions and unveiling data artifacts through influence functions.
\newblock In \emph{Proceedings of the 58th Annual Meeting of the Association for Computational Linguistics}, pages 5553--5563, Online. Association for Computational Linguistics.

\bibitem[{He et~al.(2023)He, Liang, Jiao, Zhang, Yang, Wang, Tu, Shi, and Wang}]{he2023exploring}
Zhiwei He, Tian Liang, Wenxiang Jiao, Zhuosheng Zhang, Yujiu Yang, Rui Wang, Zhaopeng Tu, Shuming Shi, and Xing Wang. 2023.
\newblock Exploring human-like translation strategy with large language models.
\newblock \emph{arXiv preprint arXiv:2305.04118}.

\bibitem[{Jiao et~al.(2023)Jiao, Wang, Huang, Wang, and Tu}]{jiao2023chatgpt}
Wenxiang Jiao, Wenxuan Wang, JT~Huang, Xing Wang, and ZP~Tu. 2023.
\newblock Is chatgpt a good translator? yes with gpt-4 as the engine.
\newblock \emph{arXiv preprint arXiv:2301.08745}.

\bibitem[{Koh and Liang(2017)}]{koh2017understanding}
Pang~Wei Koh and Percy Liang. 2017.
\newblock Understanding black-box predictions via influence functions.
\newblock In \emph{International conference on machine learning}, pages 1885--1894. PMLR.

\bibitem[{Krantz and Parks(2002)}]{krantz2002implicit}
Steven~George Krantz and Harold~R Parks. 2002.
\newblock \emph{The implicit function theorem: history, theory, and applications}.
\newblock Springer Science \& Business Media.

\bibitem[{Lam et~al.(2022)Lam, Hasler, and Hieber}]{lam-etal-2022-analyzing}
Tsz~Kin Lam, Eva Hasler, and Felix Hieber. 2022.
\newblock Analyzing the use of influence functions for instance-specific data filtering in neural machine translation.
\newblock In \emph{Proceedings of the Seventh Conference on Machine Translation (WMT)}, pages 295--309. Association for Computational Linguistics.

\bibitem[{Li et~al.(2024{\natexlab{a}})Li, Cheng, Huang, and Chen}]{li2024mt}
Jiahuan Li, Shanbo Cheng, Shujian Huang, and Jiajun Chen. 2024{\natexlab{a}}.
\newblock Mt-patcher: Selective and extendable knowledge distillation from large language models for machine translation.
\newblock \emph{arXiv preprint arXiv:2403.09522}.

\bibitem[{Li et~al.(2024{\natexlab{b}})Li, Zhou, Huang, Cheng, and Chen}]{li2024eliciting}
Jiahuan Li, Hao Zhou, Shujian Huang, Shanbo Cheng, and Jiajun Chen. 2024{\natexlab{b}}.
\newblock Eliciting the translation ability of large language models via multilingual finetuning with translation instructions.
\newblock \emph{Transactions of the Association for Computational Linguistics}, 12:576--592.

\bibitem[{Li et~al.(2023)Li, Zhang, Li, Chen, Chen, Cheng, Wang, Zhou, and Xiao}]{li2023quantity}
Ming Li, Yong Zhang, Zhitao Li, Jiuhai Chen, Lichang Chen, Ning Cheng, Jianzong Wang, Tianyi Zhou, and Jing Xiao. 2023.
\newblock From quantity to quality: Boosting llm performance with self-guided data selection for instruction tuning.
\newblock \emph{arXiv preprint arXiv:2308.12032}.

\bibitem[{Lin et~al.(2022)Lin, Mihaylov, Artetxe, Wang, Chen, Simig, Ott, Goyal, Bhosale, Du et~al.}]{lin2022few}
Xi~Victoria Lin, Todor Mihaylov, Mikel Artetxe, Tianlu Wang, Shuohui Chen, Daniel Simig, Myle Ott, Naman Goyal, Shruti Bhosale, Jingfei Du, et~al. 2022.
\newblock Few-shot learning with multilingual generative language models.
\newblock In \emph{Proceedings of the 2022 Conference on Empirical Methods in Natural Language Processing}, pages 9019--9052.

\bibitem[{Martens and Grosse(2015)}]{martens2015optimizing}
James Martens and Roger Grosse. 2015.
\newblock Optimizing neural networks with kronecker-factored approximate curvature.
\newblock In \emph{International conference on machine learning}, pages 2408--2417. PMLR.

\bibitem[{OpenAI(2023)}]{ChatGPT}
OpenAI. 2023.
\newblock \href {https://openai.com/blog/chatgpt/} {Chatgpt: Optimizing language models for dialogue.}

\bibitem[{Ouyang et~al.(2022)Ouyang, Wu, Jiang, Almeida, Wainwright, Mishkin, Zhang, Agarwal, Slama, Ray et~al.}]{ouyang2022training}
Long Ouyang, Jeffrey Wu, Xu~Jiang, Diogo Almeida, Carroll Wainwright, Pamela Mishkin, Chong Zhang, Sandhini Agarwal, Katarina Slama, Alex Ray, et~al. 2022.
\newblock Training language models to follow instructions with human feedback.
\newblock \emph{Advances in Neural Information Processing Systems}, 35:27730--27744.

\bibitem[{Papineni et~al.(2002)Papineni, Roukos, Ward, and Zhu}]{papineni2002bleu}
Kishore Papineni, Salim Roukos, Todd Ward, and Wei-Jing Zhu. 2002.
\newblock Bleu: a method for automatic evaluation of machine translation.
\newblock In \emph{Proceedings of the 40th annual meeting of the Association for Computational Linguistics}, pages 311--318.

\bibitem[{Pu et~al.(2021)Pu, Chung, Parikh, Gehrmann, and Sellam}]{pu2021learning}
Amy Pu, Hyung~Won Chung, Ankur~P Parikh, Sebastian Gehrmann, and Thibault Sellam. 2021.
\newblock Learning compact metrics for mt.
\newblock In \emph{Proceedings of EMNLP}.

\bibitem[{Raunak et~al.(2023)Raunak, Sharaf, Awadallah, and Menezes}]{raunak2023leveraging}
Vikas Raunak, Amr Sharaf, Hany~Hassan Awadallah, and Arul Menezes. 2023.
\newblock Leveraging gpt-4 for automatic translation post-editing.
\newblock \emph{arXiv preprint arXiv:2305.14878}.

\bibitem[{Rei et~al.(2020)Rei, Stewart, Farinha, and Lavie}]{rei2020comet}
Ricardo Rei, Craig Stewart, Ana~C Farinha, and Alon Lavie. 2020.
\newblock Comet: A neural framework for mt evaluation.
\newblock \emph{arXiv preprint arXiv:2009.09025}.

\bibitem[{Schioppa et~al.(2022)Schioppa, Zablotskaia, Vilar, and Sokolov}]{schioppa2022scaling}
Andrea Schioppa, Polina Zablotskaia, David Vilar, and Artem Sokolov. 2022.
\newblock Scaling up influence functions.
\newblock In \emph{Proceedings of the AAAI Conference on Artificial Intelligence}, volume~36, pages 8179--8186.

\bibitem[{Sellam et~al.(2020)Sellam, Das, and Parikh}]{sellam2020bleurt}
Thibault Sellam, Dipanjan Das, and Ankur~P Parikh. 2020.
\newblock Bleurt: Learning robust metrics for text generation.
\newblock \emph{arXiv preprint arXiv:2004.04696}.

\bibitem[{Teso et~al.(2021)Teso, Bontempelli, Giunchiglia, and Passerini}]{teso2021interactive}
Stefano Teso, Andrea Bontempelli, Fausto Giunchiglia, and Andrea Passerini. 2021.
\newblock Interactive label cleaning with example-based explanations.
\newblock \emph{Advances in Neural Information Processing Systems}, 34:12966--12977.

\bibitem[{Touvron et~al.(2023)Touvron, Martin, Stone, Albert, Almahairi, Babaei, Bashlykov, Batra, Bhargava, Bhosale et~al.}]{touvron2023llama}
Hugo Touvron, Louis Martin, Kevin Stone, Peter Albert, Amjad Almahairi, Yasmine Babaei, Nikolay Bashlykov, Soumya Batra, Prajjwal Bhargava, Shruti Bhosale, et~al. 2023.
\newblock Llama 2: Open foundation and fine-tuned chat models.
\newblock \emph{arXiv preprint arXiv:2307.09288}.

\bibitem[{Wei et~al.(2021)Wei, Bosma, Zhao, Guu, Yu, Lester, Du, Dai, and Le}]{wei2021finetuned}
Jason Wei, Maarten Bosma, Vincent~Y Zhao, Kelvin Guu, Adams~Wei Yu, Brian Lester, Nan Du, Andrew~M Dai, and Quoc~V Le. 2021.
\newblock Finetuned language models are zero-shot learners.
\newblock \emph{arXiv preprint arXiv:2109.01652}.

\bibitem[{Xia et~al.(2024)Xia, Malladi, Gururangan, Arora, and Chen}]{xia2024less}
Mengzhou Xia, Sadhika Malladi, Suchin Gururangan, Sanjeev Arora, and Danqi Chen. 2024.
\newblock Less: Selecting influential data for targeted instruction tuning.
\newblock \emph{arXiv preprint arXiv:2402.04333}.

\bibitem[{Xu et~al.(2023)Xu, Kim, Sharaf, and Awadalla}]{xu2023paradigm}
Haoran Xu, Young~Jin Kim, Amr Sharaf, and Hany~Hassan Awadalla. 2023.
\newblock A paradigm shift in machine translation: Boosting translation performance of large language models.
\newblock \emph{arXiv preprint arXiv:2309.11674}.

\bibitem[{Yang et~al.(2023)Yang, Li, Zhang, and Zong}]{yang2023bigtrans}
Wen Yang, Chong Li, Jiajun Zhang, and Chengqing Zong. 2023.
\newblock Bigtrans: Augmenting large language models with multilingual translation capability over 100 languages.
\newblock \emph{arXiv preprint arXiv:2305.18098}.

\bibitem[{Zeng et~al.(2023{\natexlab{a}})Zeng, Meng, Yin, and Zhou}]{zeng2023improving}
Jiali Zeng, Fandong Meng, Yongjing Yin, and Jie Zhou. 2023{\natexlab{a}}.
\newblock Improving machine translation with large language models: A preliminary study with cooperative decoding.
\newblock \emph{arXiv preprint arXiv:2311.02851}.

\bibitem[{Zeng et~al.(2023{\natexlab{b}})Zeng, Meng, Yin, and Zhou}]{zeng2023tim}
Jiali Zeng, Fandong Meng, Yongjing Yin, and Jie Zhou. 2023{\natexlab{b}}.
\newblock Tim: Teaching large language models to translate with comparison.
\newblock \emph{arXiv preprint arXiv:2307.04408}.

\bibitem[{Zhang et~al.(2023{\natexlab{a}})Zhang, Haddow, and Birch}]{zhang2023prompting}
Biao Zhang, Barry Haddow, and Alexandra Birch. 2023{\natexlab{a}}.
\newblock Prompting large language model for machine translation: A case study.
\newblock \emph{arXiv preprint arXiv:2301.07069}.

\bibitem[{Zhang et~al.(2023{\natexlab{b}})Zhang, Fang, Zhang, Ma, Zhou, Huang, Bu, Gui, Chen, Chen, and Feng}]{bayling}
Shaolei Zhang, Qingkai Fang, Zhuocheng Zhang, Zhengrui Ma, Yan Zhou, Langlin Huang, Mengyu Bu, Shangtong Gui, Yunji Chen, Xilin Chen, and Yang Feng. 2023{\natexlab{b}}.
\newblock Bayling: Bridging cross-lingual alignment and instruction following through interactive translation for large language models.
\newblock \emph{arXiv preprint arXiv:2306.10968}.

\bibitem[{Zhou et~al.(2023)Zhou, Liu, Xu, Iyer, Sun, Mao, Ma, Efrat, Yu, Yu et~al.}]{zhou2023lima}
Chunting Zhou, Pengfei Liu, Puxin Xu, Srini Iyer, Jiao Sun, Yuning Mao, Xuezhe Ma, Avia Efrat, Ping Yu, Lili Yu, et~al. 2023.
\newblock Lima: Less is more for alignment.
\newblock \emph{arXiv preprint arXiv:2305.11206}.

\bibitem[{Zhu et~al.(2023)Zhu, Liu, Dong, Xu, Kong, Chen, Li, and Huang}]{zhu2023multilingual}
Wenhao Zhu, Hongyi Liu, Qingxiu Dong, Jingjing Xu, Lingpeng Kong, Jiajun Chen, Lei Li, and Shujian Huang. 2023.
\newblock Multilingual machine translation with large language models: Empirical results and analysis.
\newblock \emph{arXiv preprint arXiv:2304.04675}.

\end{thebibliography}

\appendix


\section{Influence Function.} 
\label{app:if}
Influence Function (IF) was introduced to deep learning by \citep{koh2017understanding}. In the classical influence function setting, we are given the training dataset $\mathcal{D}=\{(x_i, y_i)\}_{i=1}^n$, where $x_i$ and $y_i$ are the input and label of the $i$-th training example. And the model parameter $\theta^*$ is obtained through empirical risk minimization:
\begin{align}
    \label{equ:train_obj}
    \bm{\theta}^*:&=\mathop{\arg\min}_{\bm{\theta}}\mathcal{L}(\mathcal{D}|\bm{\theta})\\ 
    &=\mathop{\arg\min}_{\bm{\theta}}\frac{1}{n}\sum_{i=1}^nL(z_i|\bm{\theta}),
\end{align}
where $z_i=(x_i, y_i)$ denotes the $i$-th input-label pair and $L$ is the loss function, e.g., cross entropy loss. Influence function measures the impact of a training example $z_m$ to the response function by up-weighting $z_m$ by $\varepsilon$ in the training objective:
\begin{equation}
    \bm{\theta}^*(\varepsilon)=\mathop{\arg\min}_{\bm{\theta}}\frac{1}{n}\sum_{i=1}^nL(z_i|\bm{\theta}) + \varepsilon L(z_m).
\end{equation}
The influence of the training example $z_m$ on the response function $\bm{\theta}^*$ is the derivative of $\bm{\theta}^*$ with respect to $\varepsilon$ at $\varepsilon=0$:
\begin{equation}
    \mathbf{\mathcal{I}_{\bm{\theta}^*}}(z_m)=\left.\frac{\rm{d}\bm{\theta^*}(\varepsilon)}{\rm{d}\varepsilon}\right|_{\varepsilon=0}.
\end{equation}
Using the Implicit Function Theorem \citep{krantz2002implicit} and first-order Taylor approximation, we define the influence
\begin{equation}
    \mathcal{I}_{\bm{\theta}^*}(z_m):=-{\bf H}^{-1}_{\bm{\theta}^*}\nabla_{\bm{\theta}|\bm{\theta}^*}L(z_m),
\end{equation}
where ${\bf H}_{\bm{\theta}^*}=\nabla_{\bm{\theta}}^2\mathcal{L}(\mathcal{D}|\bm{\theta}^*)$ is the Hessian of the training objective at $\bm{\theta}=\bm{\theta}^*$. In this paper, we are interested in the influence of the training example $z_m$ on the model behavior at the test example $z_t$, i.e., $L(z_t|\bm{\theta}^*)$. Using chain rule, we define the influence of $z_m$ on the model loss at $z_t$:
\begin{align}
    \mathcal{I}(z_m, z_t):&=\left.\frac{{\rm d}L(z_t|\bm{\theta}^*(\varepsilon))}{{\rm d}\varepsilon}\right|_{\varepsilon=0} \\
    &=-\nabla_{\bm{\theta}|\bm{\theta}^*}L(z_t)^\top{\bf H}^{-1}_{\bm{\theta}^*}\nabla_{\bm{\theta}|\bm{\theta}^*}L(z_m).
\end{align}
Since in fine tuning the model is trained with small learning rate and few steps, the model parameter at the end of training is not the minimizer of the objective in \eqref{equ:train_obj} and the Hessian might not be positive definite thus is not invertible. The following equivalence between non-converged model parameter and regularized empirical risk minimizer provides a method for modifying IF:
\begin{table*}[t!]\Huge
\renewcommand    
	\arraystretch{1.3}
\resizebox{\linewidth}{!}{
\begin{tabular}{l c c c c c c c c c c c c}
\toprule
\multirow{3}{*}{Metric} & \multicolumn{6}{c}{Zh $\Rightarrow$ En} & \multicolumn{6}{c}{De $\Rightarrow$ En}       
\\ \cmidrule(r){2-7} \cmidrule(r){8-13} 
                       & \multicolumn{3}{c}{WMT22}                  & \multicolumn{3}{c}{FLORES}                 & \multicolumn{3}{c}{WMT22} & \multicolumn{3}{c}{FLORES} \\ \cmidrule(r){2-4} \cmidrule(r){5-7} \cmidrule(r){8-10} \cmidrule(r){11-13} 
                       & {COMET} & BLEU  & BLEURT & {COMET} & BLEU  & BLEURT & COMET  & BLEU   & BLEURT  & COMET   & BLEU   & BLEURT  \\ \midrule
\textit{p value}   &    1.6e-7              & 1.6e-8   & 2.0e-6    & 7.1e-5   & 3.4e-3  & 1.6e-5  &  3.5e-2 & 1.6e-6 &  2.1e-1  &    1.7e-4 & 1.7e-3  &  1.8e-3 \\ \bottomrule
\end{tabular}
}
\caption{In this table, we present the statical analysis outcome of our experimental results in Figure \ref{fig:maingraph}. We conduct t-test on our \method compared with random baseline and present the p-value.}
\label{tb:stat-tb}

\end{table*}

\begin{lemma}\citep{goodfellow2016deep} Assuming that the model is trained via $T$ steps of Stochastic Gradient Descent (SGD) and the learning rate $\eta$ is fixed during training. Then the model parameter at the end of training $\bm{\theta}_T$ is equal to the model parameter obtained through regularized empirical risk minimization $\hat{\bm{\theta}}=\mathop{\arg\min}_{\bm{\theta}}\mathcal{L}(\mathcal{D|\bm{\theta}})+\frac{\lambda}{2}\Vert\bm{\theta}\Vert_2^2$ when the following two conditions are satisfied:
\begin{align}
    \vert \bm{I}-\eta\bm\Lambda\vert&\preceq \bm{I}, \\
    (\bm{I}-\eta\bm{\Lambda})^T&=(\bm{\Lambda}+\lambda \bm{I})^{-1}\lambda,
\end{align}
where $\bm\Lambda$ is the diagonal matrix in the eigendecomposition of 
 $~{\bf H}_{\bm{\theta}^*}={\bf Q}\bm\Lambda{\bf Q}^\top$.
\end{lemma}
Hence we approximate of the Hessian of fine tuning as ${{\bf H}_{\bm{\theta}_T}+\lambda \bm{I}}$. For fine tuning, we have:
\begin{equation}
    \mathcal{I}(z_m, z_t)=-\nabla_{\bm{\theta}}L(z_t)^\top({\bf H}_{\bm{\theta}_T}+\lambda \bm{I})^{-1}\nabla_{\bm{\theta}}L(z_m).
\end{equation}
Note that calculating the Hessian ${\bf H}_{\bm{\theta}_T}$ accurately is time-consuming. Therefore, we follow \citep{teso2021interactive} to approximate the Hessian using empirical Fisher Information Matrix (eFIM). Also, we further use the Kronecker-Factored Approximate Curvature (KFAC) \citep{martens2015optimizing} to reduce the memory usage. We implement the IF calculation based on \emph{nngeometry}\footnote{\url{https://github.com/tfjgeorge/nngeometry}}.

\section{Annotation Guidelines}
\label{app:annot}
We hire full-time translators who are fluent in both Chinese and English and translators who are fluent in both German and English. They are recruited to conduct human evaluations. The translators are shown the source sentence and two candidate translations, which are from the random subset model and \method subset model. Then the translators are required to rate on 
the translation quality from 1 to 5 with 1 the worst and 5 the best and pick the better one. All our translators are Chinese.

\section{Statistical Analysis of Experimental Results}
\label{app:stat}
We conduct statistical analysis on our experimental results as shown in Table \ref{tb:stat-tb}. 
Specifically, we use t-test to analyze the results of our \method across various dataset sizes from Figure \ref{fig:maingraph} compared with random baseline. Our experimental results have demonstrated statistical significance, as evidenced by the p-values less than 0.05 in almost all cases.

\end{document}